\documentclass[journal]{IEEEtran}
\usepackage{times}  % DO NOT CHANGE THIS
\usepackage{helvet} % DO NOT CHANGE THIS
\usepackage{courier}  % DO NOT CHANGE THIS
\usepackage[hyphens]{url}  % DO NOT CHANGE THIS
\usepackage{graphicx} % DO NOT CHANGE THIS
\usepackage{graphicx}  % DO NOT CHANGE THIS
\usepackage{times}
\usepackage{epsfig}
\usepackage{graphicx}
\usepackage{amsmath}
\usepackage{amssymb}

\usepackage{booktabs}  
\usepackage{siunitx}

\usepackage{multirow}

\usepackage{subfig}
\title{Facial Attribute Capsules for Noise Face Super Resolution}
\author{Jingwei Xin$^\dagger$, Nannan Wang$^\ddagger$, Xinrui Jiang$^\ddagger$, Jie Li$^\dagger$, Xinbo Gao$^\dagger$, Zhifeng Li $^\S $ \\
	$^\dagger$ State Key Laboratory of Integrated Services Networks, \\
	School of Electronic Engineering, Xidian University, Xi'an 710071, China \\
	$^\ddagger$ State Key Laboratory of Integrated Services Networks, \\
	School of Telecommunications Engineering, Xidian University, Xi'an 710071, China \\
	$^\S$ Tencent AI Lab, China, \\
}
 
\begin{document}

\maketitle

\begin{abstract}
Existing face super-resolution (SR) methods mainly assume the input image to be noise-free. Their performance degrades drastically when applied to real-world scenarios where the input image is always contaminated by noise. In this paper, we propose a Facial Attribute Capsules Network (FACN) to deal with the problem of high-scale super-resolution of noisy face image. Capsule is a group of neurons whose activity vector models different properties of the same entity. Inspired by the concept of capsule, we propose an integrated representation model of facial information, which named Facial Attribute Capsule (FAC). In the SR processing, we first generated a group of FACs from the input LR face, and then reconstructed the HR face from this group of FACs. Aiming to effectively improve the robustness of FAC to noise, we generate FAC in semantic, probabilistic and facial attributes manners by means of integrated learning strategy. Each FAC can be divided into two sub-capsules: Semantic Capsule (SC) and Probabilistic Capsule (PC). Them describe an explicit facial attribute in detail from two aspects of semantic representation and probability distribution. The group of FACs model an image as a combination of facial attribute information in the semantic space and probabilistic space by an attribute-disentangling way. The diverse FACs could better combine the face prior information to generate the face images with fine-grained semantic attributes. Extensive benchmark experiments show that our method achieves superior hallucination results and outperforms state-of-the-art for very low resolution (LR) noise face image super resolution.

\end{abstract}
%%%%%%%%% BODY TEXT
\section{Introduction}

Face image super resolution (SR) is a special case of general image SR, aiming to generate a High-Resolution (HR) face image from a Low-Resolution (LR) input image. It can provide more critical information for visual perception and identity analysis \cite{FR1,FR4,FR5,FR6,FR7}. However, when images are noisy and their resolutions are inadequately small (e.g. as in some real situations), there is little information available to be inferred reliably from these LR images. Very low-resolution and noisy face images not only impede human perception but also impair computer analysis.

\begin{figure}[h]
	\captionsetup{aboveskip=-0pt}
	\captionsetup{belowskip=5pt}
	\centering
	\scalebox{1}{
	\includegraphics[width=1.0\linewidth]{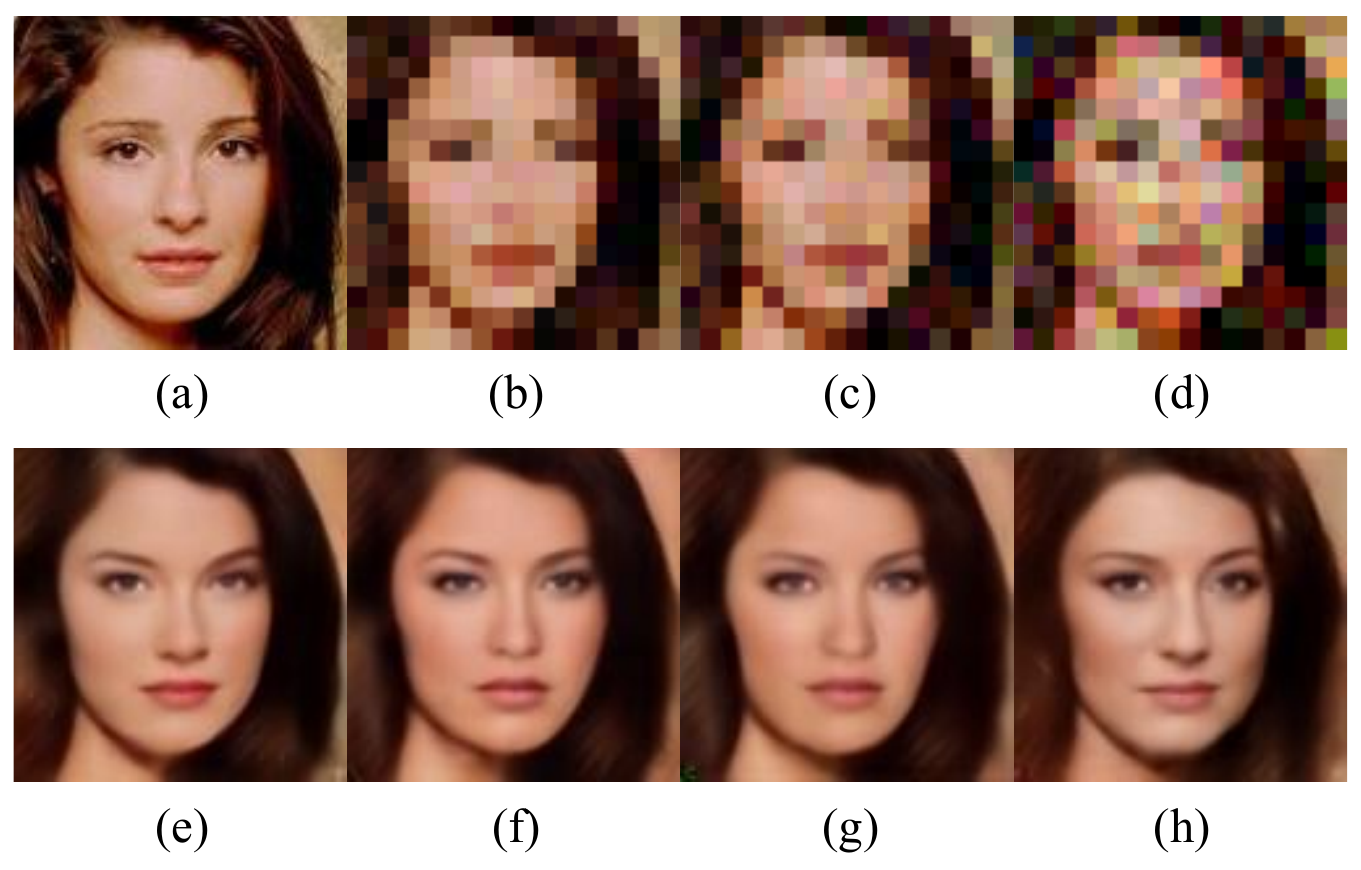}
	}
	\caption{Our face SR results on different noise levels. (a) Original HR images. (b),(c) and (d) are the blurry LR image with 5, 10 and 30 level noise. (e) Our SR result from the LR image. (f), (g) and (h) are our SR result from the (b),(c) and (d), respectively }
	\label{fig:1}
\end{figure}

Deep convolutional neural network (CNN) based Face SR methods have received significant attentions in recent years. Dong et al. \cite{SRCNN} proposed SRCNN by firstly introducing CNN to image SR, which established a nonlinear mapping from LR to HR image. Considering the feature extraction ability of deep learning, Zhou et.al \cite{BCCNN} reconstructed HR face images by combining input face images with their depth features. Face hallucination is a domain specific super-resolution problem, the prior knowledge in face images could be pivotal for face image super-resolution. Tuzel et al.\cite{GLN} proposed GLN to extract the global and local information from face images. Yu et al. \cite{URDGN} investigated GAN \cite{GAN} to create perceptually realistic HR face images. Zhu et al. \cite{CBN} proposed CBN to overcome the different face spatial configuration by dense correspondence field estimation. Tai et al. \cite{FSRNet} employed facial landmarks and parsing maps to train the network. The work also proves that the face structure prior \cite{HFP1,HFP2,HFP3,3DF} could enhance the performance of face SR. However, all of the above methods are based on image pixel level representation to super-resolve face images. Their performance degrades drastically if the input image is contaminated by noise. 

Rather than learning the deep model from the holistic appearance, the face hallucination methods, i.e.,face encoding and facial attributes, which is based on the facial semantic level representation, have been proposed. Yu et al. \cite{AEUN} introduced an encode-decode network with attribute embedding structure into face image SR problem, and proved the superiority of automEncoder in face image super resolution. The face representation feature produced by this encoding method is only a single vector, and the representation accuracy of this vector is easily reduced when the input image contaminated by noise. Thus, how to overcome the interference of noise to image reconstruction is still a problem to be solved. 

In this paper, we focus on the the problem of noise face SR and propose a Facial Attribute Capsules Network (FACN) for efficient face SR reconstruction. The image reconstruction method of FACN can be divided into two stages: At first stage, generation a group of Facial Attribute Capsules (FAC) from the input image, the second stage is the HR image reconstruction process. Each FAC could be divided into two parts: Semantic Capsule (SC) and Probability Capsule (PC). SC is a vector, where its direction represents a kind of face attribute and its norm represents the probability of the attribute exists. PC models an image as a composition of attributes in a probabilistic manner. It uses the divergence of each capsule with a prior distribution to represent the probability that an attribute exists, which maps the existing attributes into the posterior that matches the prior approximately. 

The main contributions of this work are threefold. 

(1) For face super-resolution task, we used a capsule based representation model to reconstruct HR face. Compared with the existing vector-based representation method, capsule based representation model could effectively reduce the ambiguity caused by the inherent nature of this task, especially when the target is blur and noisy.

(2) In order to effectively reduce the interference of fuzziness and noise to the coding process, we use the integrated learning strategy for reference, and carry out the facial feature coding process through semantic representation, probability distribution and attribute analysis respectively.

(3) We proposed a new capsule-based facial representation model, named FAC. Which combines the semantic representation of image and probability distribution with the rule of facial attributes. Therefore, FAC not only has strong facial representation ability of capsule based method, but also has strong noise robustness of probability distribution based method.

%-------------------------------------------------------------------------

\section{Related Work}

Face hallucination has been widely studied in recent years \cite{FH2,StructuredFH}. The classical method is mainly based on the geometric structure of the face to hallucinate HR face image. These methods can be grouped into two categories: holistic methods and part-based methods. 

Holistic methods mainly use global face models learned by PCA to recover entire HR faces. Tang et.al \cite{ET} proposed a novel approach to reconstruct HR face images by establishing a linear mapping process from LR to HR in facial subspace. Similarly, Liu \cite{FH-Theory} introduced a combination global appearance model with a local non-parametric model to enhance the facial details and achieved better performance. Kolouri et.al \cite{3} provided an efficient method to morph an HR output by optimal transport and subspace learning techniques. Due to the fact that the holistic methods are less robust to face pose variations, the input image is required to be precisely aligned. To more effectively handle various poses and expressions, a number of methods utilizing facial parts rather than entire faces have been proposed. Baker et.al \cite{4} suggested searching the best mapping between LR and HR patches can boost the capability to reconstruct high-frequency details of aligned LR face images effectively. Following this idea, \cite{sparseyang,7} blend position patches extracted from multiple aligned HR images to super-resolve aligned LR face images. Wang et. al \cite{StructuredFH} first adopted the domain knowledge of facial components in LR images and then transfers the most similar components from HR dataset to the inputs LR image. However, part-based methods are very sensitive to the local information in the input LR face image, the performance will decline sharply when noise exists.

Benefit from the learning ability of deep learning, convolutional neural network (CNN) based methods achieved state-of-the-art performance. Tuzel et al. \cite{GLN} transformed the input image into global and local feature maps by convolution and full connection. Zhu et al. \cite{BCCNN} presented an unified framework for face super-resolution and dense correspondence field estimation to recover textural details. They achieve state-of-the-art results for very low resolution inputs but fail on faces with various poses and occlusions due to the difficulty of accurate spatial prediction. Yu et al. \cite{URDGN} used the discriminant network with strong facial prior information to generate perceptually realistic HR face images. They further proposed transformative discriminative autoencoder to super-resolve unaligned, noisy and tiny LR face images \cite{TDAE}. Cao et al. \cite{AttentionFH} proposed an attention-aware face hallucination framework, which resorts to deep reinforcement learning for sequentially discovering attended patches and then performs the facial part enhancement by fully exploiting the global image interdependency. Huang et al. \cite{Wavelet} proposed a Wavelet-based CNN method, which learns to predict the LR’s corresponding series of HR’s wavelet coefficients, and utilizes them to reconstructing HR images. Chen et al. \cite{FSRNet} introduced facial landmarks and parsing maps to train the network by multi-supervision. Yu et al. \cite{AEUN} proposed an attribute embedding based coding and decoding network, which first encodes LR images with facial attributes and then super-resolves the encoded features to hallucinate LR face images. 

\begin{figure*}[t]
	\captionsetup{aboveskip=-0pt}
	\captionsetup{belowskip=5pt}
	\includegraphics[width=1.0\linewidth]{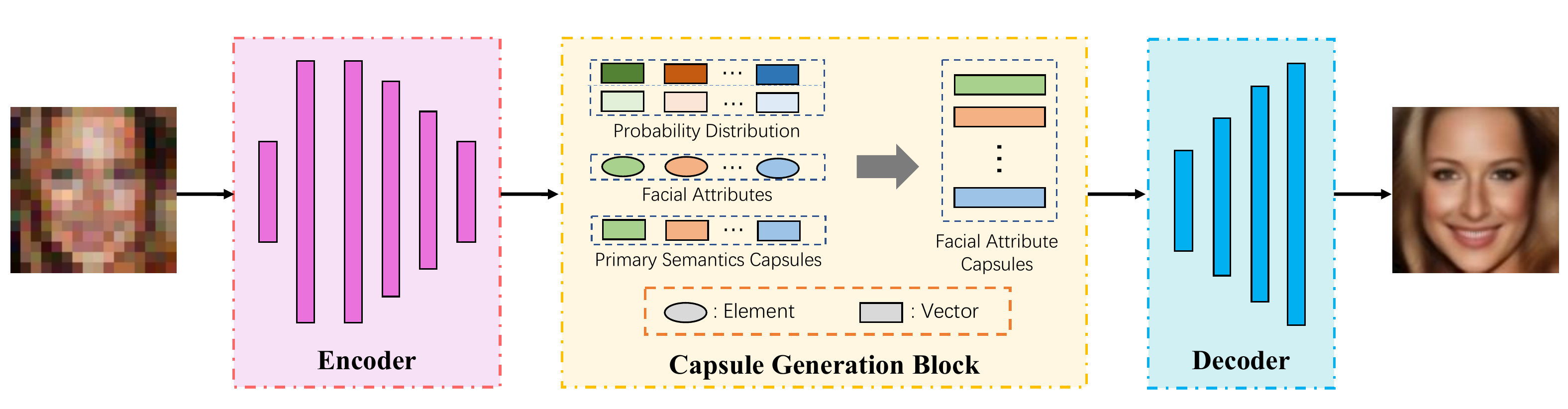}
	\caption{Pipeline of our proposed FACN model. The network consists of two parts: an Encode model to map an input image $x$ into the deep features, a Capsule Generation Block converts the features to a group of facial attribute capsules, and a Decode model to produce the output image ${\widehat y}$ from the facial attribute capsules.}
	\label{fig:2}
\end{figure*}

Hinton et al. \cite{Caps1} introduced capsules to represent properties of an image. They proposed to transform auto-encoder to learn and manipulate an image with capsules. Sabour et al. \cite{Caps2} use the length of a capsule’s activity vector to represent the probability of an entity and design an iterative routing-by-agreement mechanism to improve the performance of capsule networks. Hinton et al. \cite{Caps3} proposed a matrix version of capsules with EM routing. The proposed FAC can be seen as a new version of capsules which focus on the face image. This extends the classical capsule network to a more stable and efficient model for image generation. VAE \cite{VAE1,VAE2} is one of the most promising generative models for its theory elegancy, stable training and nice manifold representations. VAE consists of two models: an inference model to map the visible data to the latent which matches to a prior, and a generative model to synthesize the visible data from the latent code.

\section{Proposed Method: FACN}
\subsection{ Overview of FACN}
The pipeline of our proposed FCAN model is shown in Fig.\ref{fig:2}. It consists of three parts:  face SR encoder, capsule generation block and face SR decoder. Let $x$ denote the LR input image and $y$ as the final recovered HR face image. Considering the noise in the input low-resolution image may seriously interfere with the generation of facial attribute capsule, face SR encoder could be divide into two steps:,

\begin{equation}
\begin{aligned}
%\begin{array}{l}
{y_p} = P(x),  F = E({y_p}),
% \end{array}
\end{aligned}
\label{Eq:1}
\end{equation}
where $P$ denotes the nonlinear mapping from LR image $ x $ to a coarse SR image ${y_p}$, aiming to provide more sufficient facial information to the followed coding process. $E$ is the coding function and $F$ is the facial features extracted from ${y_p}$ by coding. Then the capsule generation block $G$ is utilized to generate the face attribute capsules:

\begin{equation}
\begin{aligned}
%\begin{array}{l}
Caps = G(F),
% \end{array}
\end{aligned}
\label{Eq:2}
\end{equation}
where $Caps$ is the face attribute capsules and $G$ is the function of capsule generation block. Then these capsules are fed into the face SR decoder to recover the final SR face image.

\begin{equation}
\begin{aligned}
%\begin{array}{l}
{\widehat y}=D(Caps).
% \end{array}
\end{aligned}
\label{Eq:3}
\end{equation}

Given a training set $({x^{(i)}},{y^{(i)}},{a^{(i)}})_{i = 1}^N$, where $N$ is the number of training images, $y^{(i)}$ is the ground-truth high resolution image corresponding to the low resolution image $x^{(i)}$, and $a^{(i)}$ is the corresponding ground-truth facial attribute. The loss function of the proposed FCAN is:

\begin{equation}
\begin{aligned}
L_G(\theta ) = \frac{1}{M}\sum\limits_{i = 1}^M \{ \left\| {{y^{(i)}}} - {{\widehat y}^{(i)}} \right\| + {\left\| {{y^{(i)}}} - {y_p^{(i)}} \right\|} \\
+ D_{KL} + \lambda \sum\limits_{n = 1}^N {\left\| a_{(n)}^{(i)} - {\widehat {a}_{(n)}^{(i)}} \right\|} \}, 
\end{aligned}
\label{Eq:4}
\end{equation}
where $\theta$ denotes the network parameters, $\lambda$ is the trade-off between the prior information and the prediction loss, ${{\widehat y}^{(i)}}$ and ${\widehat {a}_{(n)}^{(i)}}$ are the recovered HR image and the estimated prior attributes for the $i_{th}$ image. In addition, $D_{KL}$ is the KL-divergence which we used to train the Probability Capsule (PC). The details are described in section 3.2.2.

\subsection{Details on FACN}

\begin{figure*}[t]
	\captionsetup{aboveskip=-0pt}
	\captionsetup{belowskip=5pt}
	\centering
	\scalebox{0.9}{
	\includegraphics[width=1.0\linewidth]{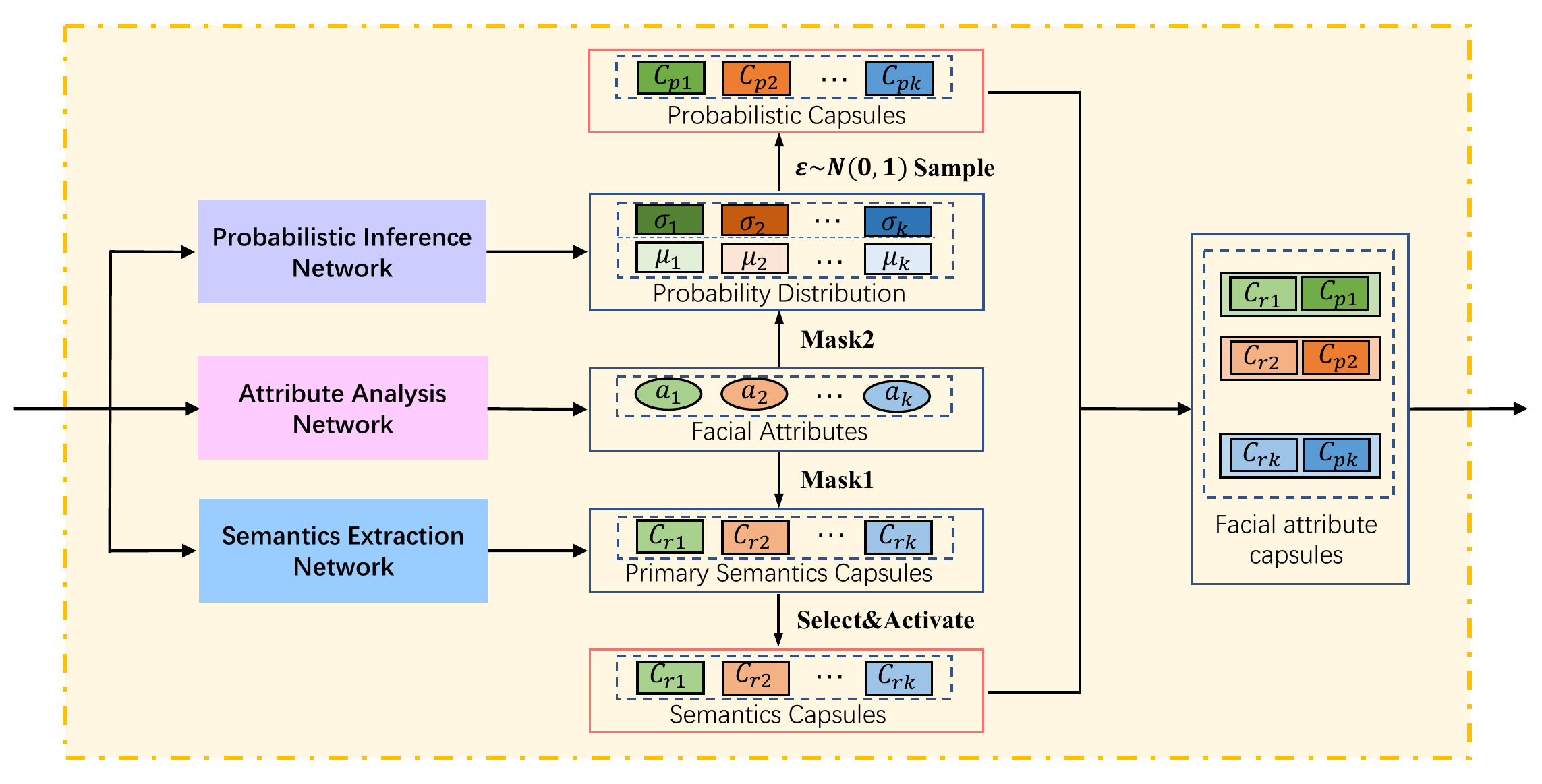}
	}
	\caption{Structure of Capsule Generation Block. The probabilistic capsules are sampled from the posterior using the reparameterization trick with a mask to indicate the present entities. Semantic capsules also be selected and activated by adding masks.}
	\label{fig:3}
\end{figure*}

We now present the details of our FACN. Where capsule generation block first generates the input facial features as representation capsules, probabilistic capsules and facial attributes. Then they are combined into the final facial attribute capsules. The structures are as shown in Fig.\ref{fig:3}.

\subsubsection{Semantic capsules and facial attributes}

The classical capsules are used to model the multiple types of objects with large differences. The shallow features extracted from the network are insufficient for the representation of multiple types of targets. Therefore, it needs the following weight matrix and dynamic routing process with huge numbers of parameters and computational complexity to obtain a group of capsules with strong feature representation ability. However, in this work, our target is to recover the HR image from the LR near-frontal face images. It is easier to capture the differences between the input images through a simple network structure, and the weight matrix and dynamic routing process are avoidable. 

Firstly, we convert the encoded features into a set of Primary Semantic Capsules (PSC) by the Semantic Extraction Network (SEN). It has three convolution layers and a fully connected layer. PSC is a vector which represent an attribute, and the length of vector represent the probability of existing attribute. The number and dimension of PSC is $k$ and $d$. The structures are shown in Fig.\ref{fig:3}. Then, for each PSC, we need to select and activate it to the SC by an attribute mask. The facial attribute is obtained by the Attribute Analysis Network (AAN), whose structure is consistent with SEN. For the features obtained by the Encoder, we have:

\begin{equation}
\begin{aligned}
%\begin{array}{l}
C_s^{pr}=S(D_f),   a_{tt}=A(D_f),
% \end{array}
\end{aligned}
\label{Eq:5}
\end{equation}
where $D_f$ is the output of the encoder, $A$ and $S$ are the function of AAN and SEN. $C_s^{pr}$ and $a_{tt}$ are the PSC and facial attribute. Then, the task of capsule selection and activate process is finished by adding attribute mask. 
\begin{equation}
\begin{aligned}
%\begin{array}{l}
{C_{\rm{s}}} = {a_{tt}} \frac{{C_{s}^{pr}}}{{\left\| {C_{s}^{pr}} \right\|}},
% \end{array}
\end{aligned}
\label{Eq:6}
\end{equation} 
where $C_s$ is the final Semantic Capsules. The latter part of the formula represents the unit vector of the $C_s$. These attributes are used to select capsules and update the their length. The facial attribute $a_{tt}$ is a vector with $k$ dimensions. The value of each dimension ranges from 0 to 1. It is inefficient to extract high frequency information from noise and low resolution image by consuming more computing and storage resources. For efficiency, we set the capsules number $k = 64$ and dimension $d = 4$ in all our experiments.

\subsubsection{Probabilistic capsules}

The semantic representation ability of capsules will be precipitous decline when the task is an ill-posed problem (for example, image super resolution, denoise and deblur). This is also an unavoidable phenomenon in the low-level task. As we all know, the variational model based methods have strong noise robustness, which could efficiently realize the nonlinear mapping between the different probability distributions. In this work, we also construct a Probabilistic Inference Network (PIN) and design a probabilistic capsules. Which follows a known prior distribution to reconstruct the image. 
\begin{equation}
\begin{aligned}
%\begin{array}{l}
{\mu},{\sigma ^2}=P(D_f),
% \end{array}
\end{aligned}
\label{Eq:7}
\end{equation}
where $\mu$ and ${\sigma ^2}$ are mean and variance. These two variables are the output vectors of the encoder. $P$ is the function of PIN. Following VAEs \cite{VAE1}, we select the KL-divergence as the metric to indicate the degree how two distributions match to each other. The KL-divergence of each capsule with the prior distribution represents the probability that a capsule’s entity exists, i.e., the capsule corresponding to the existing entity has a small KL-divergence with the prior while those corresponding to the non-existing entities have large KL-divergences with the prior distribution. Let the prior be the centred isotropic multivariate Gaussian $N(0; 1)$ and the probabilistic capsules $N(\mu; \sigma ^2)$. Then the KL-divergence term, given $N$ data samples, can be computed as:

\begin{equation}
\begin{aligned}
%\begin{array}{l}
{D_{KL}} = \frac{1}{2}\sum\limits_{i = 1}^k {(1 + \log (\sigma _i^2) - \mu _i^2 - \sigma _i^2)},
% \end{array}
\end{aligned}
\label{Eq:8}
\end{equation}

More approximate probability distribution can make the nonlinear mapping process of features more efficient and more convenient. We further utilize the facial attributes to apdate the probability distributions of PC and prior information.
\begin{equation}
\begin{aligned}
%\begin{array}{l}
{\widehat \mu}=\mu + a_{tt},
% \end{array}
\end{aligned}
\label{Eq:9}
\end{equation}
We utilize the facial attributes to adjust the mean value $\mu$ in PC. Here PC has the same dimensions as SC, i.e., the mean and variance of PC has $k$ dimensions. Then, the capsules are sampled using the reparameterization trick:
\begin{equation}
\begin{aligned}
%\begin{array}{l}
{C_{p}} = \hat \mu  + \varepsilon  \odot \sigma,
% \end{array}
\end{aligned}
\label{Eq:10}
\end{equation}
where $\varepsilon \sim N(0,1)$ is a random vector and ${\odot}$ means the element-wise multiplication.

\subsubsection{Encoder and decoder}

Our encoder could be divide into two parts: a pre-SR part and an encoding part. Firstly, we used a pre-SR network to roughly recover a coarse HR image and then code the coarse HR image, which has two $3\times3$ convolutional layers and three 3 residual blocks. The rationale behind this is that it is non-trivial to estimate facial attribute capsules from the input image. Using the coarse SR network could provide more useful imformation for the followed capsule generation process. The pre-SR part starts with a $3\times3$ convolution followed by 3 residual blocks. Another $3\times3$ convolutional layer is used to reconstruct the coarse HR image. Then, let $k3$ denote that the convolution kernel size is 3 and $s1$ denotes that the stride is 1. The encoding part architecture is: $k3s2$, $k3s1$, $k3s2$, $k3s1$, $k3s2$, $k3s1$, $k3s2$, $k3s1$, $k3s2$, $k3s2$, $k3s1$. For the decoder, it recovers the high-resolution face image directly from the FAC. The decoding part architecture is start with a fully connected layer followed by six up-sampling convolution layers. Finally, a $3\times3$ convolution layer is used to reconstruct the HR face image. All convolution channels is set to 64.

\subsubsection{Discrimination module}

GAN-based methods \cite{GAN}, formulated as a two-player game between a generator and a discriminator, have been widely used for image generation \cite{SRGAN}. Because of its prominent features (such as symmetry of contour, similarity of components), we propose to incorporate GAN into our framework. The key idea is to use a discriminant network to distinguish the super-resolved images and the real high-resolution images and use a generative network to train the SR network to deceive the discriminator.

Our discriminant network consists of eight convolution layers and two full connection layers. The objective function of the adversarial network $D$ is expressed as

\begin{equation}
\begin{aligned}
%\begin{array}{l}
{L_D}(G,D) = {\rm E}[\log D(\widehat y,x)] \\
+ {\rm E}[\log (1 - D(G(x),x))],
% \end{array}
\end{aligned}
\label{Eq:11}
\end{equation}
where $E$ is the expectation of the log probability distribution and $D$ is the generative model. Apart from the adversarial loss $L_D$, we further introduce a perceptual loss \cite{SRGAN} using high-level feature maps (i.e., features from ‘relu5-3’ layer) of our discriminant network to help assess perceptually relevant characteristics,

\begin{equation}
\begin{aligned}
%\begin{array}{l}
{L_P} = {\left\| {\phi (y) - \phi ({\widehat y})} \right\|^2},
\end{aligned}
\label{Eq:12}
\end{equation}
where $\phi$ denotes the fixed pre-trained VGG model, and maps the images $y/{\widehat y}$ to the feature space. In this way, the loss function of our generative model could be formulated as

\begin{equation}
\begin{aligned}
\arg \mathop {\min }\limits_G \mathop {\max }\limits_D {L_G}(\theta ) + \gamma_D {L_D}(G,D) + \gamma_P {L_P},
\end{aligned}
\label{Eq:13}
\end{equation}
where $\gamma$ is the trade-off between the discriminant loss and the aforementioned FACN loss.

\begin{figure}[b]
	\captionsetup{belowskip=-15pt}
	\centering
	\scalebox{1}{
	\includegraphics[width=1.0\linewidth]{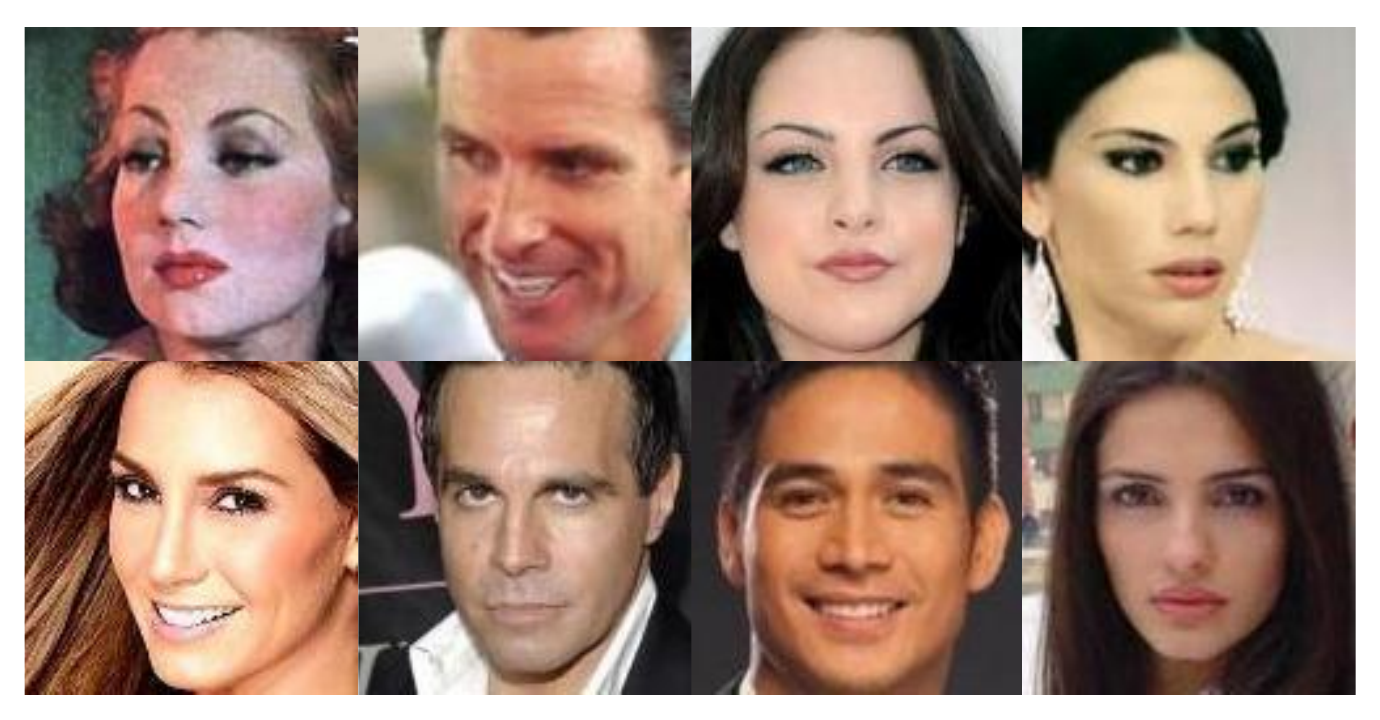}
	}
	\caption{Training examples of CelebA.}
	\label{fig:4}
\end{figure}

\begin{figure*}[h]
	\captionsetup{aboveskip=-0pt}
	\captionsetup{belowskip=-10pt}
	\centering
	\scalebox{0.9}{
	\includegraphics[width=1.0\linewidth]{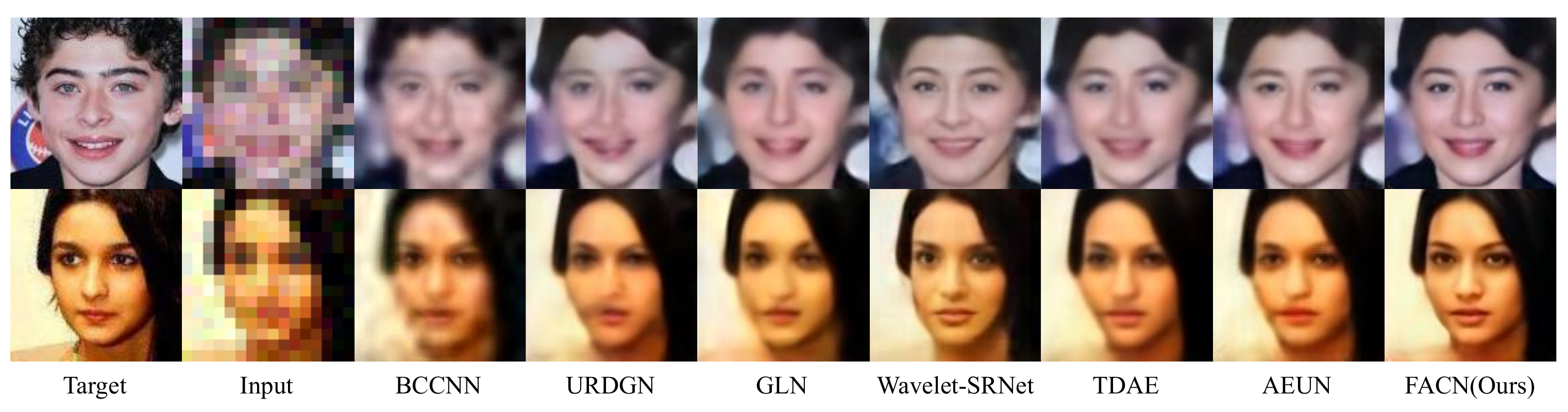}
	}
	\caption{Visual evaluation with $BicN$ degradation model.}
	\label{fig:6}
\end{figure*}

\begin{figure*}[t]
	\captionsetup{aboveskip=-0pt}
	\captionsetup{belowskip=-10pt}
	\centering
	\scalebox{0.9}{
		\includegraphics[width=1.0\linewidth]{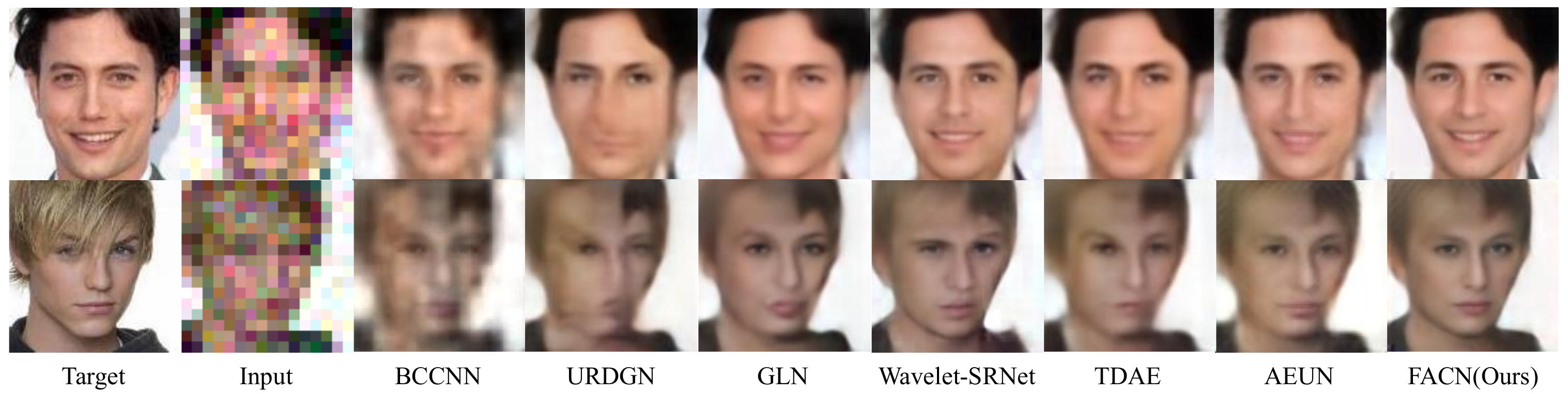}
	}
	\caption{Visual evaluation with $BBicN$ degradation model}
	\label{fig:7}
\end{figure*}

\begin{table*}[h]
	\captionsetup{aboveskip=5pt}
	\captionsetup{belowskip=-10pt}
	\small
	\centering
	\scalebox{1.0}{
		\begin{tabular}{p{2cm}p{1.0cm}p{1.0cm}p{1.0cm}p{1.0cm}p{1.0cm}p{1.0cm}p{1.0cm}p{1.0cm}p{1.0cm}}
			\toprule
			\multirow{2}{*}{Methods} & \multicolumn{3}{c}{$Bic$} & \multicolumn{3}{c}{$BicN$} &\multicolumn{3}{c}{$BBicN$} \\
			% \cline{3-6} \cline{7-10} \cline{11-14}
			
			& PSRN & SSIM & IFC & PSRN & SSIM & IFC & PSRN & SSIM & IFC \\
			\midrule
			Bicubic & 23.98 & 0.6505 & 0.6588 & 23.13 & 0.6088 & 0.4497 & 21.09 & 0.5329 & 0.3457 \\
			
			BCCNN 	& 25.29 & 0.7135 & 0.9524 & 23.94 & 0.6615 & 0.6677 & 22.21 & 0.6154 & 0.5453 \\
			
			GLN & 26.04 & 0.7427 & 1.0783 & 24.01 & 0.6718 & 0.7179 & 22.45 & 0.6365 & 0.5758  \\
			
			URDGN & 24.54 & 0.6785 & 0.6981 & 23.80 & 0.6444 & 0.5502 & 21.01 & 0.5482 & 0.3650 \\
			
			Wavelet-SRNet & 24.43 & 0.6891 & 0.7835 & 23.95 & 0.6768 & 0.7270 & 22.48 & 0.6428 & 0.6035 \\
			
			TDAE & 26.29 & 0.7411 & 1.1523 & 24.16 & 0.6778 & 0.7321 & 22.81 & 0.6511 & 0.6211 \\
			
			AEUN & 26.37 & 0.7477 & 1.1605 & 24.24 & 0.6801 & 0.7535 & 22.83 & 0.6514 & 0.6254 \\
			\midrule
			FCAN (Ours)  & {\bf 26.79} & {\bf 0.7684} & {\bf 1.2515} & {\bf 24.61} & {\bf 0.7009} & {\bf 0.8060} & {\bf 23.14} & {\bf 0.6714} & {\bf 0.6775} \\
			\bottomrule
		\end{tabular}
	}
	\caption{Benchmark results with different degradation model.}
	\label{tab:1}
\end{table*}

%------------------------------------------------------------------------

\section{Experiments}

\subsection{Implementation}

\textbf{Dataset} 
We conduct experiments on celebA dataset \cite{Celeba}. We use the first 36000 images for training, and the following 1000 images for testing. We coarsely crop the training images according to their face regions and resize to $128\times128$ without any pre-alignment operation. Examples from the training data set are shown in Fig.\ref{fig:4}. Here we use color images for training as SRGAN does \cite{SRGAN}. In addition, each image has 40 attribute annotations. We exclude some attributes which are not necessary such as hair or skin colors. As a result, we choose 18 attributes, such as gender, age, and beard information from 40 attributes, and use these attributes to supervise the top 18 elements of the output of AAN. Other attributes are regarded as potential facial attributes and let them learn freely in an unrestricted state. 

\textbf{Degradation models} 
In order to fully demonstrate the effectiveness of our proposed FACN for noise and blurring, we use three degradation models to simulate LR images. The first one is bicubic downsampling by adopting the Matlab function imresize with the option bicubic (denote as $Bic$ for short). We use $Bic$ model to simulate LR images with scaling factor $8$. The second one is to downsample with scaling factor $8$, and then add Gaussian noise with noise level $10$ \cite{RDN} (denote as $BicN$ for short), where the noise level $ n $  means a standard deviation $ n $ in a pixel intensity range of [0, 255]. We further produce LR image in a more challenging way. We first blur HR image by Gaussian kernel of size $7\times7$ with standard deviation $1.6$, and bicubic downsample HR image with scaling factor $8$, then add Gaussian noise with noise level $30$ (denote as $BBicN$ for short).

\textbf{Training setting} 
We initialize the convolutional layers as the same as He \textit{et al.} \cite{Ini}. All convolutional layers are followed by LeakyReLU \cite{Leak} with a negative slope of 0.2. We implement our model using the pytorch environment, and optimize our network by Adam with back propagation. The momentum parameter is set to 0.5, weight decay is set to $1 \times {10^{{\rm{ - }}4}}$, and the initial learning rate is set to $3 \times {10^{{\rm{ - }}4}}$ and being divided a half every 20 epochs.The batch size is set to 16. We empirically set $\lambda = 1$, ${\gamma_P} = 0.01$ and ${\gamma_D} = 0.01$. Training a basic FACN on celebA dataset generally takes 10 hours with one Titan X Pascal GPU. For assessing the quality of SR results, we employ two objective image quality assessment metrics: Peak Signal to Noise Ratio (PSNR) and structural similarity (SSIM) \cite{SSIM}. All metrics are performed on the Y-channel (YCbCr color space) of super-resolved images.

\subsection{Comparisons with State-of-the-Art Methods}

We compare our proposed FCAN with state-of-the-art SR methods, including BCCNN \cite{BCCNN}, GLN \cite{GLN}, Wavelet-SRNet \cite{Wavelet}, TDAE \cite{TDAE} and AEUN \cite{AEUN}. For fair comparison, we train all models with the same training set. In order to achieve higher performance, we only train the image generation model for the GAN-based methods, but the entire GAN network for qualitative comparisons.

Tab.\ref{tab:1} summarizes quantitative results on the Celeba datasets. Our FACN significantly outperforms state-of-the-arts in both PSNR and SSIM. We follow the same experimental setting on handling occlued face as Wavelet-SRNet \cite{Wavelet} and directly import the $16\times16$ test examples for super-resolving $128\times128$ HR images. Benefiting from a more efficient integrated representation approach of facial information, our method produces relatively sharper edges and shapes, while other methods may give more blurry results.

Then, we compared our FACN with state-of-the-art methods in a noise environment. As shown in Fig.\ref{fig:6} and Fig.\ref{fig:7}. Under the effect of noise, the performance of all methods has been reduced, but our method can still have a more clear face, especially the eyes and nose. AEUN can be seen as an improvement version by introducing the face attribute information to TADE. Thus the individual components of face image are generated more clearly. In addition, our method has very strong noise robustness in qualitative results. As shown in fig.\ref{fig:1}, the visual quality of our reconstructed face image does not changed significantly with the increase of noise level.

In order to corroborate the real benefit of the proposal, we further perform the face verification experiments via the Arcface \cite{deng2019arcface}. We constructed 1000 positive sample pairs and 9000 negative sample pairs based on the SR results (BBicN) from each method. Results are shown in Tab. \ref{tab:3}. It can be seen that our reconstruction results have better identity retention property.

\begin{table}[h]
	\captionsetup{belowskip=-5pt}
	\captionsetup{aboveskip=5pt}
	\small
	\centering
	\begin{tabular}{p{2.0cm}p{1.5cm}p{1.5cm}p{1.5cm}}
		\toprule
		Methods & Performance & Methods & Performance  \\
		\midrule
		Bicubic & 0.8058 & BCCNN & 0.8570   \\
		URDGN & 0.8212 & GLN & 0.8580  \\
		Wavelet-SRNet & 0.8820 & TDAE & 0.8680 \\
		AEUN & 0.8694 & FACN(Ours) & {\bf 0.8922}   \\
		\bottomrule
	\end{tabular}
	\caption{Face recognition evaluation on the $BBicN$ degradation SR results from each method.}
	\label{tab:3}
\end{table}

\subsection{Ablation Study}

\textbf{Effect of FAC} We conduct ablation study on the effects of our facial attribute capsules. Since our network has the similar network structure as classical capsule based autoencoder \cite{Caps2}, we clearly show how the performance improves with semantic capsules, probabilistic capsules and classical capsules. We conduct 4 experiments to estimate the semantic capsules, probabilistic capsules, and FAC, respectively. Specifically, by removing the probabilistic capsules from our basic FACN, the remaining parts constitute the first network, named ‘BasicNet v1’. The second network, named ‘BasicNet v2’, has the same structure as ‘BasicNet v1’ except that the removing part is the semantic capsules. The third network ‘BasicNet v3’ repalce the Capsule Generation Block (CGB) by the method of \cite{Caps2}, which generates the capsules by a weight matrix and dynamic routing process. In this part, we only analyze the quality of different types of capsules. For fairly comparison, the differences among the four networks are only limited to the part of CGB, the encoders and decoders have same structure.

\begin{figure}[t]
	\captionsetup{aboveskip=-0pt}
	\captionsetup{belowskip=0pt}
	\centering
	\scalebox{1}{
	\includegraphics[width=1.0\linewidth]{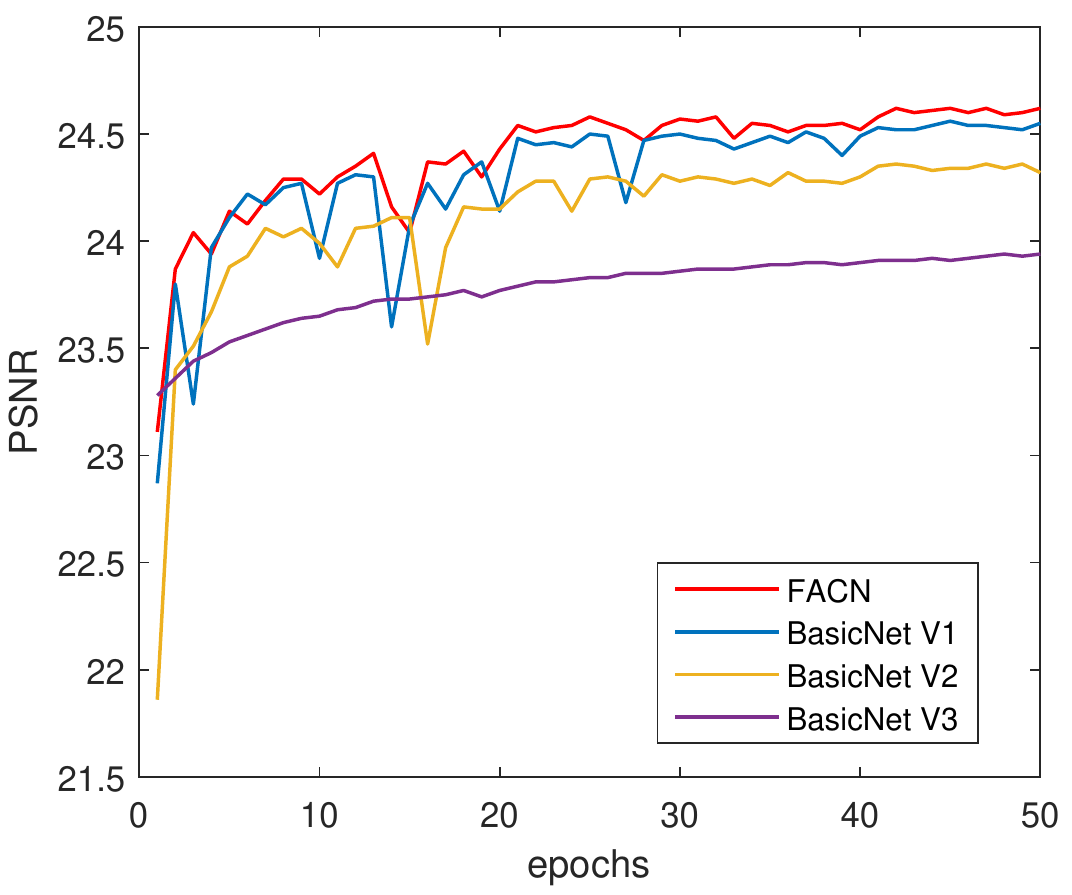}
	}
	\caption{Ablation study on effects of facial attribute capsules with $BicN$ degradation model.}
	\label{fig:8}
\end{figure}

Fig.\ref{fig:8} shows the results of different network structures. It can be seen that: (1) Compared to the other capsules, classic capsules (BasicNet v3) are not suitable for face image super resolution because of its limited ability to faces representation. (2) Semantic capsules (BasicNet v1) are qualified for face image super-resolution, and the probabilistic capsules (BasicNet v2) worked alone has inferior performance and blur results. (3) The model using both capsules (semantic and probabilistic) achieves the best performance, which indicates richer prior information brings more improvement.

Actually, the phenomenon of gradient explosion always exist in the training of classic capsule network. We think this is caused by the shallow features of the network are difficult to fully represent the input image. When the classic capsule network adopted our encoder and decoder, this phenomenon has been significantly alleviated. In spite of this, from the results after network convergence, it can be seen that there is still a big gap between the classical capsules and our semantic capsules. This also indicates that the ambiguity is significantly reduced by imposing explicit semantic information into the capsules.

\textbf{Effect of capsules numbers and dimensions} In this part, we conduct ablation study on the effects of the number and dimension of FAC. We first study the effect of the capsules numbers $k$ in the capsules generation block. Specifically, we test $k = 18/32/64/128$, and the PSNR results are shown in the Tab.\ref{tab:2}, respectively. Due to the number of supervised attributes is 18,  the minimum of the number of capsules is 18. We can find that during the increase in the number of capsules from 18 to 64, performance improves faster. But when the number increased from 64 to 128, the performance improved slowly.

\begin{table}[h]
	\captionsetup{belowskip=0pt}
	\captionsetup{aboveskip=5pt}
	\small
	\centering
	\begin{tabular}{p{1.0cm}p{1.0cm}p{1.0cm}p{1.0cm}p{1.0cm}}
		\toprule
		k & 18 & 32 & 64 & 128 \\
		\midrule
		PSNR & 24.25 & 24.49 & 24.61 & 24.68  \\
		\midrule
		d & 2 & 4 & 8 & 16 \\
		\midrule
		PSNR & 24.45 & 24.61 & 24.69 & 24.75  \\
		\bottomrule
	\end{tabular}
	\caption{Ablation study on effects of capsules numbers and dimensions with $BicN$ degradation model.}
	\label{tab:2}
\end{table}

We also study the effect of the capsules dimension $d$ and the results shown in the Tab.\ref{tab:2}. Since using more dimensions leads to a wider structure, the representation ability of the FAC grows, and hence better performance. Finally, for a compromise between network performance and computational complexity, we choose $k=64$ and $d=4$ in this work.

\section{Conclusion}

In this paper, we propose a novel image super resolution network which is named Facial Attribute Capsule Network (FACN). FACN could provide a more comprehensive face representation mode (the Facial Attribute Capsule), and show the obvious advantages in the super-resolution reconstruction of noise face images. In order to improve the robustness of face representation model to noise and blur, FACN encodes the face images by combining semantic representation and probability distribution. Extensive benchmark experiments show that FACN significantly outperforms the state-of-the-arts. This compact object representation mode could be widely applicabled in practice of other machine vision problems such as inpainting, compression artifact removal and even recognition.

\section{Acknowledgement}
This work was supported in part by the National Natural Science Foundation of China under Grant Grant 61922066, Grant 61876142, Grant 61671339, Grant 61772402, Grant U1605252, Grant 61432014, in part by the National Key Research and Development Program of China under Grant 2016QY01W0200 and Grant 2018AAA0103202, in part by the National High-Level Talents Special Support Program of China under Grant CS31117200001, in part by the Fundamental Research Funds for the Central Universities under Grant JB190117, in part by the Xidian University-Intellifusion Joint Innovation Laboratory of Artificial Intelligence, in part by the Innovation Fund of Xidian University.

\bibliographystyle{IEEEtran}
\bibliography{FACN}

\end{document}